\PassOptionsToPackage{table,dvipsnames}{xcolor}
\PassOptionsToPackage{sort&compress}{natbib}
\PassOptionsToPackage{pdfusetitle}{hyperref}
\documentclass[11pt,a4paper]{thuc3i}
\usepackage{natbib}
\renewcommand{\cite}{\citep}
\usepackage{placeins}
\usepackage{tcolorbox}
\tcbuselibrary{breakable,listings,skins}
\usepackage{booktabs}
\usepackage[table]{xcolor}
\definecolor{gain}{HTML}{1A7F37}
\definecolor{loss}{HTML}{B42318}
\definecolor{mleC}{HTML}{A9BCE6}
\definecolor{mleT}{HTML}{DDE5F5}
\definecolor{kbC}{HTML}{A783CC}
\definecolor{kbT}{HTML}{E6DAF2}
\definecolor{avgC}{HTML}{AFD3A8}
\definecolor{avgT}{HTML}{DFEEDB}

\graphicspath{{figures/}}
\definecolor{gain}{HTML}{1A7F37}
\definecolor{loss}{HTML}{B42318}
\newcommand{\tabup}[2]{\shortstack[r]{#1\\[-1pt]{\scriptsize\textcolor{gain}{+#2}}}}
\newcommand{\tabdn}[2]{\shortstack[r]{#1\\[-1pt]{\scriptsize\textcolor{loss}{$-$#2}}}}

\definecolor{promptblueframe}{RGB}{45, 95, 145}
\definecolor{promptbluetitle}{RGB}{45, 95, 145}
\newtcblisting{promptbox}[1]{
  enhanced, listing only, title={#1},
  colback=white, colframe=promptblueframe, colbacktitle=promptbluetitle,
  coltitle=white, fonttitle=\small\bfseries, titlerule=0pt,
  listing options={
    basicstyle=\ttfamily\scriptsize, breaklines=true, breakatwhitespace=false,
    columns=fullflexible, keepspaces=true, showstringspaces=false, tabsize=2
  },
  boxrule=0.6pt, arc=1.5mm,
  left=1mm, right=1mm, top=1mm, bottom=1mm,
  before skip=5pt, after skip=5pt
}
\newcommand{\mcell}[1]{\begin{tabular}[t]{@{}l@{}}#1\end{tabular}}

\title{Demystifying the Autoresearch Harness:\\ How Context and Skills Shape Self-Evolving Agents}

\author{%
  \mbox{Kelvin Kiu-Wai Tam\textsuperscript{1,*}},
  \mbox{Newt Hue-Nam K. Nguyen\textsuperscript{1,*}},
  \mbox{Tianshi Zheng\textsuperscript{1,*,\textdagger}},
  \mbox{Yunxiang Mo\textsuperscript{1}},
  \mbox{Weihan Li\textsuperscript{2}},
  \mbox{Zhongwei Xie\textsuperscript{1}},
  \mbox{Chiu Yeung Tang\textsuperscript{1}},
  \mbox{Jui-Yu Ho\textsuperscript{3}},
  \mbox{Yangqiu Song\textsuperscript{1,\textdagger}},
  \mbox{Ginny Y. Wong\textsuperscript{4}},
  \mbox{Simon See\textsuperscript{4}}
  \par\vskip4pt
  {\normalfont\fontsize{8.5}{11}\selectfont
    \textsuperscript{1}The Hong Kong University of Science and Technology\quad
    \textsuperscript{2}The University of Tokyo\\
    \textsuperscript{3}National Taiwan University\quad
    \textsuperscript{4}NVIDIA AI Technology Center\par\vskip2pt
    \textsuperscript{*}Equal contribution.\quad
    \textsuperscript{\textdagger}Corresponding author.\par}
}
\hypersetup{pdfauthor={Kelvin Kiu-Wai Tam, Newt Hue-Nam K. Nguyen, Tianshi Zheng,
  Yunxiang Mo, Weihan Li, Zhongwei Xie, Chiu Yeung Tang, Jui-Yu Ho,
  Yangqiu Song, Ginny Wong, Simon See}}
\setheadertext{Demystifying the Autoresearch Harness}
\begin{abstract}
Autoresearch agents, which autonomously design, run, and refine experiments, are accelerating scientific progress, particularly in AI research. However, existing agents rely on harnesses that combine multiple mechanisms, making it difficult to determine how individual components contribute to performance. We systematically study harness design for long-horizon, self-evolving autoresearch agents along two dimensions: context management, which determines what an agent retains from its past attempts, and skill management, which governs how it stores and reuses lessons from those attempts. Across MLE-Bench and KernelBench with \texttt{gpt-oss-120b}, \texttt{gpt-6-luna}, and \texttt{gpt-6-sol}, we isolate the effects of individual harness design choices by evaluating each against a recency-truncation baseline within a shared agent framework. For context management, we find that preserving individually accessible records of past attempts is more effective than maintaining a rolling summary, with context folding performing best across all six settings and improving upon the baseline by up to 19.2\%.
For skill management, selectively discarding persistently unused lessons is more effective on average than consolidating them or promoting them to standing rules. Building on these findings, we combine context folding and skill deletion in MLEvolver, which achieves a 72.7\% medal rate on MLE-Bench Lite and a 94\% $\mathrm{fast}_1$ score on KernelBench Level~3, substantially outperforming AIDE and OpenEvolve under the same inference budget. Overall, our study distills design principles for preserving experimental history and curating reusable lessons in autoresearch harnesses, paving the way for agents that turn accumulated experience into sustained progress.
\end{abstract}

\begin{document}
\maketitle\par
\section{Introduction}

Autoresearch agents can automate the cycle at the heart of empirical research: given a task and a grader, they propose an idea, implement it, run the experiment, read the result, and decide what to try next. They are becoming increasingly important in AI for AI (AI4AI), where improving existing AI systems directly accelerates subsequent research. Recent autoresearch agents demonstrate this across diverse domains: on MLE-Bench \citep{chan2025mlebench}, AIDE \citep{jiang2025aide}, and ML-Master \citep{liu2025mlmaster} iteratively train and tune competitive machine learning pipelines. On KernelBench \citep{ouyang2025kernelbench}, Kevin \citep{baronio2026kevin} learns to write and iteratively refine fast CUDA kernels, and AlphaEvolve \citep{novikov2025alphaevolve} has even sped up the training of the language models it is built on. What is common among these workflows is their long horizons operation. The agent attempts the same task tens or hundreds of times, receives an evaluation score after each attempt, and iteratively refines its solution.
\begin{figure}[t]
  \centering
  \includegraphics[width=.86\linewidth,height=.38\textheight,keepaspectratio]{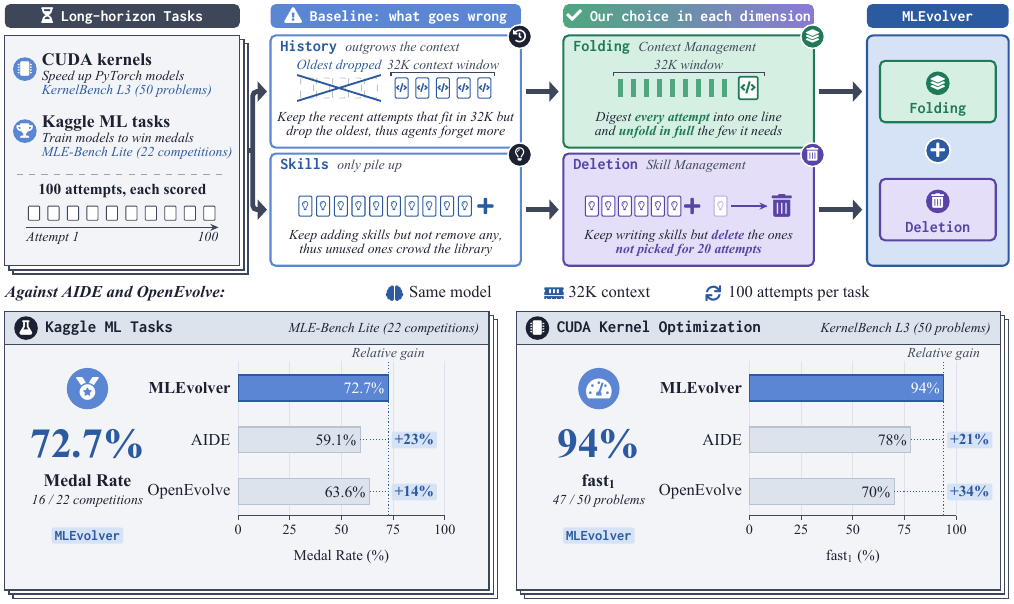}
  \caption{We evaluate context- and skill-management mechanisms for long-horizon AI autoresearch on Kaggle competitions in MLE-Bench and CUDA kernel optimization tasks in KernelBench. MLEvolver combines context folding and skill deletion, the strongest mechanisms in their respective dimensions. MLEvolver outperforms AIDE and OpenEvolve on both benchmarks, achieving a 72.7\% medal rate across 22 MLE-Bench Lite competitions and 94\% on $\mathrm{fast}_1$ across 50 KernelBench Level~3 problems.}
  \label{fig:overview}
\end{figure}

Over such horizons, an agent's behavior depends not only on its language model, but also on the harness that manages the memory it accumulates about its own past. This memory takes two forms, each under a different kind of pressure. The first is the \textit{history} of past attempts. It grows with every attempt, and soon exceeds the context window, thereby leading to truncation \citep{packer2023memgpt}. The second is a library of \textit{skills}: natural language lessons distilled from earlier attempts and retrieved when a similar situation recurs \citep{shinn2023reflexion, zhao2024expel}. The library can grow without limit, but it can become redundant and noisy, which makes the right lessons harder to retrieve. What an agent keeps, condenses, or discards in each form of memory is therefore a central design decision. 

\begin{figure}[t]
  \centering
  \includegraphics[width=\linewidth]{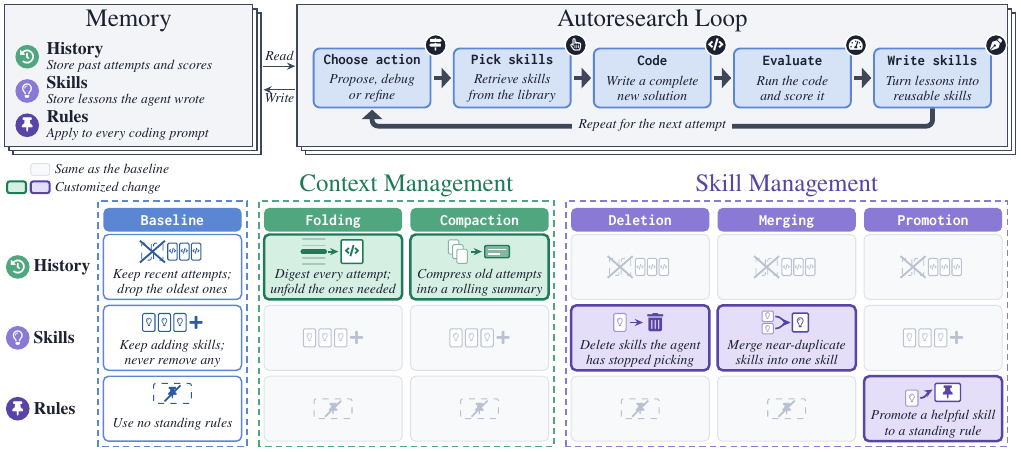}
  \caption{\textsc{Autoresearch Loop} : 
\textbf{Top:} the autoresearch loop that every
configuration repeats, and the three memory
components it reads and writes.
\textbf{Bottom:} the six configurations. Each variant changes exactly one component of the baseline (blue): context management (green) changes the
history, and skill management (purple) changes
the skills or the rules. Colored cells mark the
change; faded cells are the same as the
baseline.}
  \label{fig:method}
\end{figure}

Many mechanisms have been proposed for managing this memory. Agents page information in and out of context \citep{packer2023memgpt}, reflect on and summarize past experience \citep{park2023generative, shinn2023reflexion}, and extract, revise, and reuse insights or workflows \citep{zhao2024expel}. However, these mechanisms are typically evaluated within multi-component systems where differences in search policies, prompt structure, and tool use confound cross-system comparisons. Even when individual studies ablate their own designs, memory mechanisms are rarely evaluated head-to-head under a shared, unified harness. As a result, it remains unclear what each mechanism contributes on its own.  %

We address these questions through a controlled, bottom-up study of memory harness design in \textit{iterative self-improvement}, where an autoresearch agent repeatedly attempts the same task and receives empirical feedback after each attempt. This setting strains both memory tiers early: the execution history quickly overflows the context window, while the skill library accumulates redundant lessons. Starting from a minimal recency-truncation baseline, we isolate five memory mechanisms within an otherwise identical agent by changing exactly one component at a time (Figure~\ref{fig:method}). Two manage working context: \textit{folding} condenses each attempt into an expandable digest, and \textit{compaction} aggregates older history into a running summary. Three govern the skill library: \textit{deletion} prunes unused lessons, \textit{merging} consolidates semantic duplicates, and \textit{promotion} elevates proven skills into persistent standing rules.

We evaluate three model backbones, \texttt{gpt-oss-120b}, \texttt{gpt-6-luna}, and \texttt{gpt-6-sol}, on two AI4AI benchmarks whose feedback differs sharply. On MLE-Bench~\citep{chan2025mlebench}, the agent writes machine learning solutions for Kaggle competitions spanning tabular, text, image, and audio data, and receives a score after each attempt. We report a gold-normalized score (GNS), which is 0 for a chance submission and 100 at the gold medal threshold. On KernelBench \citep{ouyang2025kernelbench}, the agent writes CUDA kernels for PyTorch programs, from single operators to full models, and receives compilation, correctness, and timing feedback. We report the geometric-mean speedup over PyTorch Eager. Within each task, all configurations start from the same seed solution, giving them a common starting point for measuring improvement over iterations.

For context management, folding achieves the highest performance among all evaluated mechanisms in all six model--benchmark settings, improving MLE-Bench GNS by up to 7.9\% and KernelBench geometric-mean speedup by up to 19.2\% over recency truncation. Our context analysis links this advantage to folding's preservation of a separate digest for every attempt, which exposes more of the agent's history within the context budget. Compaction compresses older attempts into a single rolling summary, leaving progressively less detail about each experiment as the history grows. Folding keeps individual attempts accessible throughout the run, helping the agent draw on earlier successes and failures when choosing its next experiment.

For skill management, deletion yields the largest average improvement at 4.1\% over the recency-truncation baseline. These results favor the removal of persistently unused lessons over merging similar skills or promoting them to standing rules. Our skill analysis also reveals consistent differences in retrieval behavior across the three backbones. Across configurations, \texttt{gpt-oss-120b} selects 6.4--7.1 skills per attempt, compared with 1.2--1.9 for \texttt{gpt-6-luna} and 0.5--0.6 for \texttt{gpt-6-sol}. This ordering persists across the baseline and all three skill-management mechanisms, indicating that the backbone strongly influences how much stored experience enters each attempt.

Guided by these findings, we build MLEvolver by combining context folding and skill deletion, the best-performing mechanisms on average in their respective dimensions. We compare MLEvolver with AIDE and OpenEvolve using \texttt{gpt-6-sol} and a shared cap of 100 iterations per run. As shown in Figure~\ref{fig:overview}, MLEvolver earns medals in 16 of 22 MLE-Bench Lite competitions, achieving a 72.7\% medal rate. On KernelBench Level~3, it produces correct kernels faster than PyTorch Eager on 47 of 50 problems, reaching 94\% on $\mathrm{fast}_1$. Averaged across these two benchmarks, MLEvolver achieves relative gains of 22\% over AIDE and 24\% over OpenEvolve.

Our contributions are as follows:
\begin{itemize}
    \item A controlled study of harness design for long-horizon, self-evolving autoresearch agents: six configurations within one agent that isolate five context- and skill-management mechanisms against a recency-truncation baseline.
    \item An evaluation across three model backbones and two AI4AI benchmarks, with analyses of the effectiveness of per-attempt digests in folding, the dual benefits of skill retention and active management, and differences in skill retrieval behavior across backbones.
    \item MLEvolver, a harness that combines context folding and skill deletion and substantially outperforms AIDE and OpenEvolve on MLE-Bench Lite and KernelBench Level~3.
\end{itemize}

\section{Related Work}
\paragraph{Harness Design and Memory in Autoresearch.}

Autoresearch harnesses orchestrate iterative discovery by structuring how language models store, retrieve, and build upon past attempts. Existing systems couple search logic with bespoke memory stores: FunSearch \citep{romeraparedes2024funsearch} and AlphaEvolve \citep{novikov2025alphaevolve} maintain program databases, AIDE \citep{jiang2025aide} manages candidate trees, and ML-Master \citep{liu2025mlmaster} integrates a selectively scoped execution log into its harness. Broader agent harnesses also develop targeted mechanisms to mitigate context bounds. These harnesses page state in and out of prompt buffers \citep{packer2023memgpt}, summarize completed sub-goals \citep{hu-etal-2025-hiagent}, train models to fold execution trajectories \citep{ye2026agentfold, sun2026contextfolding}, or mask older observations to match summarization at reduced token cost \citep{lindenbauer2025complexity}. For procedural skills, prior frameworks accumulate verbal reflections \citep{shinn2023reflexion}, distilled insights \citep{zhao2024expel}, or synthesized experience replays \citep{liu-etal-2025-contextual}. However, uncurated skill catalogs degrade retrieval accuracy as they expand \citep{li2026dynamicskills}. Recent harnesses address this bloat through heuristic forgetting \citep{zhong2023memorybank}, pruning low-utility records \citep{xiong-etal-2026-memory}, or injecting curated operational playbooks into working context \citep{chen2024automanual, zhang2026ace}. Unlike prior frameworks designed for single trajectories, cross-task transfer, or brief retries, autoresearch harnesses must sustain refinement across long horizons of scored attempts. In our study, we compare five context- and skill-management mechanisms within a shared harness, varying one component at a time.

\paragraph{AI4AI Benchmarks.} 
A growing set of benchmarks employs agents to automate the development of AI systems themselves, a direction that ML-Master \citep{liu2025mlmaster}, MLAgentBench \citep{huang2024mlagentbench} and MLE-Bench \citep{chan2025mlebench} task agents with improving machine learning models on research problems and Kaggle competitions. KernelBench \citep{ouyang2025kernelbench} and TritonBench \citep{li-etal-2025-tritonbench} focus on writing correct and fast GPU kernels. On RE-Bench \citep{wijk2025rebench}, agents outperform human experts under short time budgets but gain less from longer ones, so the effectiveness of making use of a long run of attempts remains an open problem for autoresearch agents. We use MLE-Bench and KernelBench, where every attempt is scored, so each mechanism's effect can be tracked throughout a run.

\section{Memory Mechanisms}
In this section, we introduce the iterative self-evolution procedure and detail the memory mechanisms across two aspects: context management (\S\ref{cm}) and skill management (\S\ref{sg}). Figure~\ref{fig:method} illustrates one iteration and how each memory mechanism interacts with the agent.

\subsection{The Iterative Self-Evolution Loop}

Each task is solved through a sequence of scored evolution iterations. At iteration $t$, the agent first selects an action from three operators: \texttt{propose\_new}, which develops a new approach; \texttt{refine\_current}, which modifies the incumbent solution; and \texttt{debug\_current}, which addresses a failure in the incumbent. A skill picker then selects reusable skills from the task-specific skill library. The coder receives the task, standing rules, selected skills, and the representation of the previous attempt history, and produces the next solution. The benchmark evaluator executes the solution and returns task-specific feedback. Finally, a skill writer records lessons from the iteration. Skill creation remains an agent decision and any iteration can request zero or more skills.

We use the same loop across all experimental configurations. Let $H_t=(a_0,a_1,\ldots,a_{t-1})$ denote the attempt history before iteration $t$, where $a_0$ is a common seed solution and each subsequent $a_i$ contains the generated solution, evaluation feedback, and associated reasoning. The coder receives
\begin{equation}
C_t = [T,R_t,K_t,\Phi(H_t)],
\qquad
|C_t|\leq B,
\end{equation}
where $T$ is the task description, $R_t$ is the set of standing rules, $K_t$ is the subset of the task's skill library selected by the picker, $\Phi$ is the configuration-specific history policy, and $B=32{,}000$ tokens is the fixed context budget.

\subsection{Baseline}
The baseline uses recency truncation. Let $B_H$ denote the portion of the context budget available to the attempt history after accounting for the other prompt components. The baseline retains the longest suffix that fits:
\begin{equation} \begin{gathered} (a_j,\ldots,a_{t-1}),\\ j=\min\{k: |(a_k,\ldots,a_{t-1})|\le B_H\}. \end{gathered} \end{equation}

When the history exceeds the available budget, the oldest attempts are discarded first. A discarded attempt cannot be retrieved later. The skill library remains independent of this truncation and continues to accumulate skills that can be selected by the picker.

\subsection{Context Management}
\label{cm}

Context management operates on the growing attempt history $H_t$. We compare recency truncation with two mechanisms that preserve information from older attempts in compressed form: folding and compaction.

\paragraph{Folding.}
Folding represents every previous attempt using a compact diagnostic digest while allowing selected attempts to be unfolded in full. A summarizer $\sigma$ maps each attempt $a_i$ to a digest $\hat{a}_i=\sigma(a_i)$. At iteration $t$, the agent receives the sequence of digests and selects a subset $U_t$ of attempts for full expansion:
\begin{equation} \Phi_{\mathrm{fold}}(H_t)= (\hat{a}_0,\ldots,\hat{a}_{t-1}) \oplus (a_i)_{i\in U_t}. \end{equation}

Unfolded attempts are admitted according to their selected priority until the history budget $B_H$ is reached. Every attempt remains addressable throughout the run, so historical information can be recovered based on relevance.

\paragraph{Compaction.}
Compaction maintains a running summary $m$ together with a tail $\tau$ of recent attempts. When the coder context reaches the configured window, the history since the previous compaction is divided into a recent tail $\tau$ and an older head $h$. The head is incorporated into the running summary:
\begin{equation}
m \leftarrow S(m,h),
\qquad
\Phi_{\mathrm{comp}}(H_t)=m\oplus\tau.
\end{equation}

The summary follows a fixed structure containing the task objective, important details, and current work state. Unlike folding, compaction does not preserve individual attempts that can later be selected and expanded.

\subsection{Skill Management}
\label{sg}

While context management governs the attempt history, the agent also maintains a second memory tier: a task-local skill library $L_t$. Each task initializes an empty library, and an active skill reaches the coder only when selected by the skill picker. To prevent catalog bloat and control retrieval overhead, we evaluate three management mechanisms: pruning via deletion, consolidation via merging, and elevation via promotion.

\paragraph{Deletion.}
Deletion removes skills that remain unused for a sustained period. Let $u_k(t)$ denote the number of consecutive iterations for which skill $k$ has not been selected. We set the deletion threshold to $\tau_d=5$:
\begin{equation}
\left\{ 
k\in L_t:
u_k(t)<\tau_d
\right\}.
\end{equation}

A skill is therefore removed after five consecutive iterations without selection. Deletion changes only the live skill catalog and does not modify the attempt history or standing rules.

\paragraph{Merging.}
Merging consolidates newly written skills that are semantically close to an existing skill. Each skill $k$ has an embedding $e(k)$. For a newly written skill $k'$, its nearest live neighbor is \begin{equation} k^*=\arg\max_{k\in L_t} \cos(e(k'),e(k)). \end{equation} If $\cos(e(k'),e(k^*))\ge\theta$ with $\theta=0.8$, the two skills are merged. Otherwise, $k'$ is added as a new skill. Merging therefore reduces redundancy in the skill library.

\paragraph{Promotion.}
Promotion converts selected reusable skills into standing rules. Unlike ordinary skills, standing rules are included in every subsequent coder prompt and do not require retrieval by the skill picker.

A promotion agent examines the skill library and usage statistics. For each skill, the promoter observes how often it was selected and how often it was present during an iteration that produced a new best result. It then nominates a subset $N_t=\pi(L_t)$ for promotion. A nominee must pass an admissibility check $A$ requiring that it contributed to at least one improvement. Accepted nominees are added to the standing-rule set $R_t$ and removed from the ordinary skill library, yielding $R_{t+1}=R_t\cup A(N_t)$ and $L_{t+1}=L_t\setminus A(N_t)$.

\newpage
\section{Experiments}
\label{sec:experiments}
\subsection{Experimental Setup}
\label{sec:experimental_setup}
We evaluate the five memory mechanisms and baseline on two machine learning benchmarks. MLE-Bench evaluates agents on Kaggle machine-learning competitions \citep{chan2025mlebench}, while KernelBench evaluates agents' ability to generate fast and correct GPU kernels for machine-learning workloads \citep{ouyang2025kernelbench}. We use three model backbones, \texttt{gpt-oss-120b}~\citep{openai2025gptoss120bgptoss20bmodel}, \texttt{gpt-6-luna}~\citep{openai2026gpt6luna} and \texttt{gpt-6-sol}~\citep{openai2026gpt6sol}. For merging, \texttt{Qwen3-Embedding-0.6B}~\cite{zhang2025qwen3embeddingadvancingtext} is used to generate the skill embeddings.

Each task has an independent skill library, which is reset before the task begins. All configurations for a given task start from the same seed solution that is curated to be deliberately weak. The memory policy determines how the attempt history and skill information occupy the context.

\paragraph{MLE-Bench.} 
\label{sec:mle-bench}
We evaluate on 10 competitions from MLE-Bench \citep{chan2025mlebench}, comprising six low-complexity and four medium-complexity competitions. The selected competitions and benchmark-specific modifications are detailed in Appendix~\ref{mle-modification}. 

Each configuration runs 100 iterations per competition. In each iteration, the agent's solution trains on the competition's training data and writes a submission, which is then graded. The agent then receives its score under the competition's metric, together with any terminal output produced during execution. Because the metrics differ in scale and direction, we report a gold-normalized score (GNS), analogous to the human-normalized score in reinforcement learning \citep{mnih2015human}:
\begin{equation}
  \mathrm{GNS}_i = 100 \cdot \frac{s_i - s_i^{0}}{s_i^{\mathrm{gold}} - s_i^{0}},
\end{equation}
where $s_i$ is a run's best score on competition $i$, $s_i^{0}$ the score of a do-nothing submission, and $s_i^{\mathrm{gold}}$ the gold medal threshold on MLE-Bench's leaderboard for that competition \citep{chan2025mlebench}. We aggregate over the $N = 10$ competitions with weights $w_i$ of 1 and 2 for low- and medium-complexity competitions:
\begin{equation}
  \mathrm{GNS} = \frac{\sum_{i=1}^{N} w_i\, \mathrm{GNS}_i}{\sum_{i=1}^{N} w_i}.
\end{equation}

\paragraph{KernelBench.}
We evaluate 50 KernelBench problems across three difficulty levels \citep{ouyang2025kernelbench}: 20 Level-1 individual operators, 20 Level-2 fused operator chains, and 10 Level-3 model architectures. Appendix~\ref{app:kb-sampling} describes how the evaluation subset is constructed.

Each configuration runs for 50 refinement iterations per problem. At each iteration, the generated CUDA kernel is compiled and evaluated against the PyTorch Eager reference implementation on randomly generated inputs. The evaluation runner applies static and runtime safeguards against reward hacking, including checks for prohibited code patterns and protection against reusing reference outputs. Implementations that violate these checks are invalidated. Appendix~\ref{app:kb-environment} describes the correctness and timing procedures, while Appendix~\ref{app:kb-antihack} details the reward-hacking safeguards.

We report the geometric-mean speedup over PyTorch Eager,
consistent with recent evaluations of agent-generated GPU kernels \citep{dai2026cudaagentlargescaleagentic,younesian2026agentkernelarenageneralizationawarebenchmarkinggpu}. The geometric mean provides a continuous summary of multiplicative performance gains across the 50 tasks and is less sensitive than the arithmetic mean to exceptionally large speedups. 

\subsection{Main Results}

\label{sec:main-results}

\label{sec:main-results}

\begin{table}[!htbp]
\centering
\small
\begin{tabular*}{\linewidth}{@{}l@{\extracolsep{\fill}}ccccccc@{}}
\toprule
 & \multicolumn{3}{c}{\textbf{MLE-Bench} (GNS $\uparrow$)} & \multicolumn{3}{c}{\textbf{KernelBench} (GM speedup $\uparrow$)} & \\
\cmidrule(lr){2-4}\cmidrule(lr){5-7}
\textbf{Method} & gpt-oss-120b & gpt-6-luna & gpt-6-sol
                & gpt-oss-120b & gpt-6-luna & gpt-6-sol & \textbf{Avg.\ $\Delta$\%} \\
\midrule
Baseline & 70.5 & 86.1 & 93.2 & 1.17 & 1.46 & 2.16 & -- \\
\addlinespace[3pt]
\multicolumn{8}{@{}l}{\textit{Context management}} \\
\quad Folding    & \tabup{\textbf{76.1}}{7.9} & \tabup{\textbf{89.7}}{4.2} & \tabup{\textbf{94.7}}{1.6}
                 & \tabup{\textbf{1.29}}{10.3} & \tabup{\textbf{1.74}}{19.2} & \tabup{\textbf{2.34}}{8.3} & \textbf{+8.6} \\
\quad Compaction & \tabup{71.6}{1.6} & \tabup{86.4}{0.3} & \tabdn{91.6}{1.7}
                 & \tabdn{1.16}{0.9} & \tabup{1.47}{0.7} & \tabup{2.20}{1.9} & +0.3 \\
\addlinespace[3pt]
\multicolumn{8}{@{}l}{\textit{Skill management}} \\
\quad Deletion   & \tabup{\underline{73.8}}{4.7} & \tabup{\underline{87.6}}{1.7} & \tabup{94.1}{1.0}
                 & \tabup{\underline{1.25}}{6.8} & \tabup{\underline{1.54}}{5.5} & \tabup{\underline{2.26}}{4.6} & \underline{+4.1} \\
\quad Merging    & \tabup{73.7}{4.5} & \tabup{87.1}{1.2} & \tabup{94.0}{0.9}
                 & \tabup{1.22}{4.3} & \tabup{1.47}{0.7} & \tabup{2.19}{1.4} & +2.2 \\
\quad Promotion  & \tabup{73.6}{4.4} & \tabup{87.0}{1.0} & \tabup{\underline{94.3}}{1.2}
                 & \tabup{1.24}{6.0} & \tabup{1.48}{1.4} & \tabup{2.24}{3.7} & +3.0 \\
\bottomrule
\end{tabular*}
\caption{Main results. MLE-Bench reports the gold-normalized
score (GNS); KernelBench reports the geometric-mean speedup
over PyTorch Eager. Small numbers give the relative change (\%)
from the displayed baseline score in the same column.
Avg.\ $\Delta$\% is the mean of the six displayed changes.
Best per column in \textbf{bold}, second best
\underline{underlined}.}
\label{tab:main-results}
\end{table}

\paragraph{Folding performs best across all model backbones and both benchmarks.} 
Table~\ref{tab:main-results} reports results across 10 MLE-Bench competitions and 50 KernelBench tasks. Folding achieves the highest performance among all evaluated mechanisms in all six settings, with an average relative improvement of 8.6\% over the baseline. Its gains range from 1.6\% to 7.9\% on MLE-Bench and 8.3\% to 19.2\% on KernelBench. With \texttt{gpt-oss-120b}, folding raises GNS from 70.5 to 76.1 and geometric-mean speedup from $1.17\times$ to $1.29\times$. With \texttt{gpt-6-sol}, it raises GNS from 93.2 to 94.7 and geometric-mean speedup from $2.16\times$ to $2.34\times$. These improvements span the lowest- and highest-performing backbones in our study. The context analysis in Section~\ref{sec:analysis-folding} examines how retaining individually accessible records of earlier attempts contributes to this consistent advantage.

\paragraph{Skill management consistently improves performance, with deletion performing best on average.}
Deletion, merging, and promotion improve on the baseline for all models and benchmarks. Deletion yields the largest average relative gain at 4.1\%, followed by promotion at 3.0\% and merging at 2.2\%. Averaged across the three mechanisms and both benchmarks, the gains are 5.1\% for \texttt{gpt-oss-120b}, 1.9\% for \texttt{gpt-6-luna}, and 2.1\% for \texttt{gpt-6-sol}. Compared with \texttt{gpt-oss-120b}, \texttt{gpt-6-luna} gains 3.2 percentage points less on each benchmark. For \texttt{gpt-6-sol}, the gap is 3.5 points on MLE-Bench and 2.5 points on KernelBench. Section~\ref{sec:analysis-skill-utilization} analyzes this difference by investigating how the three models interact with the skill library.

\paragraph{Compaction remains near the recency-truncation baseline.}
Compaction yields little improvement in final task performance on average, with a mean relative gain of 0.3\%. On MLE-Bench, it raises GNS from 70.5 to 71.6 for \texttt{gpt-oss-120b}, but lowers it from 93.2 to 91.6 for \texttt{gpt-6-sol}. On KernelBench, it increases geometric-mean speedup from $2.16\times$ to $2.20\times$ for \texttt{gpt-6-sol}, but decreases it from $1.17\times$ to $1.16\times$ for \texttt{gpt-oss-120b}. These mixed results show that although compaction reduces the context occupied by earlier attempts, it cannot produce a consistent performance gain over the baseline. Section~\ref{sec:analysis-folding} examines this performance--cost trade-off and compares how much information various context management mechanisms retain from earlier attempts.

\paragraph{Most gains emerge early, but later iterations continue to refine the solutions.}
Figure~\ref{fig:evolution} shows performance over iterations for all three model backbones. Averaged across the six configurations and three backbones, KernelBench geometric-mean speedup increases by 65.2\% in the first half and by a further 7.4\% in the second half. On MLE-Bench, GNS increases by an average of 41.6 points in the first half and by another 2.9 points in the second half. Thus, most improvement occurs during the first half of the run, while every configuration continues to make measurable progress during later iterations. In addition, folding leads at the midpoint in five of the six model--benchmark settings and finishes first in all six. This indicates that its advantage emerges early and persists through subsequent refinement.
\section{Analysis}
\label{sec:analysis}

\subsection{How Context Affects Agent Performance}
\label{sec:analysis-folding}
We quantify how often an attempt advances the incumbent solution using the \textit{new-best rate}: the share of attempts whose score beats every earlier attempt of the same run. We report it for the first 10\% of the iterations and for the rest.

\begin{figure}[!htbp]
    \centering
    \includegraphics[width=\linewidth,height=.31\textheight,keepaspectratio]{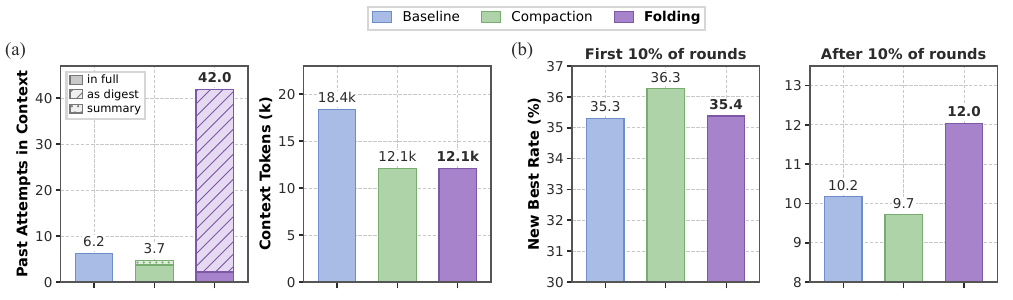}
    \caption{Context usage and progress under different context-management mechanisms. (a)~Average number of past attempts and average context tokens in the coder's prompt. (b)~New-best rate for the three mechanisms.}
    \label{fig:context}
\end{figure}

The three context management mechanisms achieve similar new-best rates during the first 10\% of iterations. As shown in Figure~\ref{fig:context}b, folding reaches 35.4\%, compared with 36.3\% for compaction and 35.3\% for the baseline. After the first 10\% of iterations, when more past attempts have accumulated, folding achieves the highest average new-best rate at 12.0\%, compared with 10.2\% for the baseline and 9.7\% for compaction. Folding therefore discovers improvements about 18\% more frequently than the baseline and 24\% more frequently than compaction during this later stage.

Figure~\ref{fig:context}a shows how folding expands access to experimental history. The baseline retains an average of 6.2 recent attempts in full and discards earlier history. In contrast, folding condenses each attempt into a compact digest of the attempted strategy and its outcome, retaining records of 42.0 past attempts on average while allowing the agent to recover full records when needed. It achieves this broader coverage with approximately 12.1k context tokens on average, compared with 18.4k for the baseline. Access to these prior successes and failures allows the agent to target each new attempt more effectively. 

Compaction also uses approximately 12.1k tokens on average, but it achieves a lower later-stage new-best rate. It merges all older attempts into one summary of bounded length, which cannot keep pace with a growing history. Together, these results indicate that reducing context usage alone is insufficient to sustain improvement. The representation must also preserve useful access to accumulated experimental history.

\subsection{Separating Skill-Library and Management Gains}
\label{sec:analysis-gain-decomposition}
\begin{figure}[!htbp]
  \centering
  \includegraphics[width=\linewidth,height=.31\textheight,keepaspectratio]{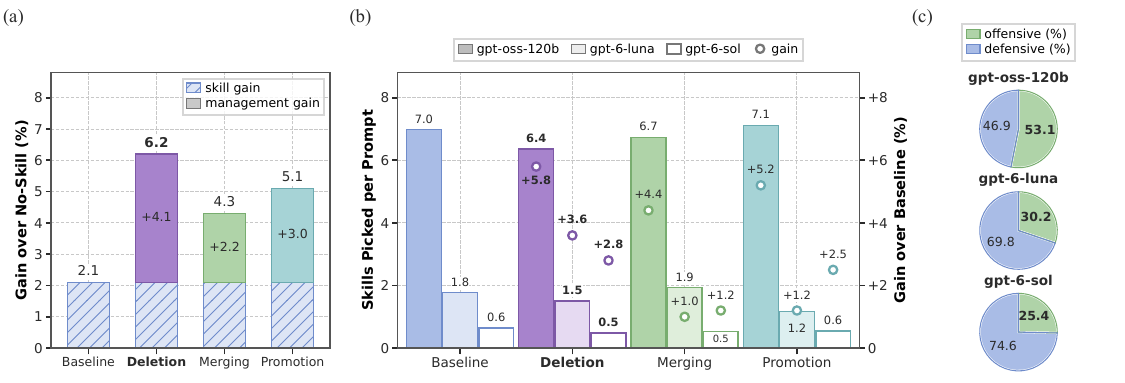}
  \caption{Performance and skill retrieval under different skill-management mechanisms. (a)~Gains over the no-skill agent. Each mechanism adds its management gain (solid segment) to the baseline’s library gain (hatched segment). (b)~Skills picked per prompt (bars, left axis) and gain over the same model’s baseline (dots, right axis), averaged across both benchmarks. (c)~Share of offensive and defensive skills among the selected skills, averaged over MLE-Bench and KernelBench.}
  \label{fig:skill-management-related}
\end{figure}

To separate the benefit of maintaining a skill library from the additional benefits of managing it, we compare each mechanism with a No-Skill ablation in which the skill writer and picker are disabled. Figure~\ref{fig:skill-management-related}a reports relative performance changes computed separately for each model--benchmark setting and then averaged across settings.

Introducing an unmanaged skill library improves average performance by 2.1\% over No-Skill, showing that recording and retrieving lessons benefits iterative refinement. Managing those lessons provides further gains over the unmanaged baseline, with deletion yielding the largest average improvement at 4.1\%, followed by promotion at 3.0\% and merging at 2.2\%.

Deletion uses the model backbone's recent skill selections as a signal of relevance, pruning persistently unselected lessons as the run progresses. This also reduces the number of entries the picker must consider. As shown in Figure~\ref{fig:skill-library}, on MLE-Bench with \texttt{gpt-oss-120b}, deletion leaves 31 skills at the end of the run, compared with 312 for the unmanaged baseline and 330 for promotion.

Merging and promotion use different criteria, which explains their smaller average gains. Merging consolidates semantically similar skills, but can weaken specific applicability conditions or implementation details. Promotion keeps lessons from earlier attempts as standing rules, even when the solution evolves and their original conditions no longer apply. Appendix~\ref{app:skill-management-cases} illustrates these issues with case studies from both benchmarks.

\subsection{Model Differences in Skill Retrieval and Orientation}
\label{sec:analysis-skill-utilization}
The different gains from skill management coincide with differences in how the three model backbones use the skill library. Figure~\ref{fig:skill-management-related}b shows that \texttt{gpt-oss-120b} retrieves 6.4--7.1 skills per prompt across configurations, whereas \texttt{gpt-6-luna} retrieves 1.2--1.9 and \texttt{gpt-6-sol} retrieves 0.5--0.6. This ordering holds across the baseline and all three skill-management mechanisms, suggesting consistent differences in the models' skill selection behavior.

The models also differ in the orientation of the knowledge drawn from the skill library. We classify a skill as \textit{offensive} when it encourages a new implementation, optimization, or search direction intended to improve the score, and as \textit{defensive} when it primarily protects the current solution by preserving correctness or avoiding known failures. Full classification criteria are provided in Appendix~\ref{app:offense-defense}. As shown in Figure~\ref{fig:skill-management-related}c, 53.1\% of the skills retrieved by \texttt{gpt-oss-120b} are offensive, compared with 30.2\% for \texttt{gpt-6-luna} and 25.4\% for \texttt{gpt-6-sol}.

Together, these observations suggest that the two proprietary backbones, \texttt{gpt-6-luna} and \texttt{gpt-6-sol}, use the skill library more conservatively. They draw on fewer stored experiences and more often select knowledge aimed at protecting existing performance. The open-weight backbone, \texttt{gpt-oss-120b}, draws on a broader set of experiences with a more even balance between protective guidance and active optimization. This difference offers one explanation for why skill management provides larger gains for \texttt{gpt-oss-120b}.
\section{MLEvolver: Combining the Best Choice in Each Dimension}
\label{sec:mlevolver}

Context folding and skill deletion deliver the largest average gains among the context-management and skill-management mechanisms, respectively. We combine them in MLEvolver and compare the resulting harness with AIDE and OpenEvolve on 22 competitions in MLE-Bench Lite and 50 problems in KernelBench Level~3. We track how each system improves over 100 iterations and compare its final performance.

\begin{figure}[!htbp]
  \centering
  \includegraphics[width=\linewidth,keepaspectratio]{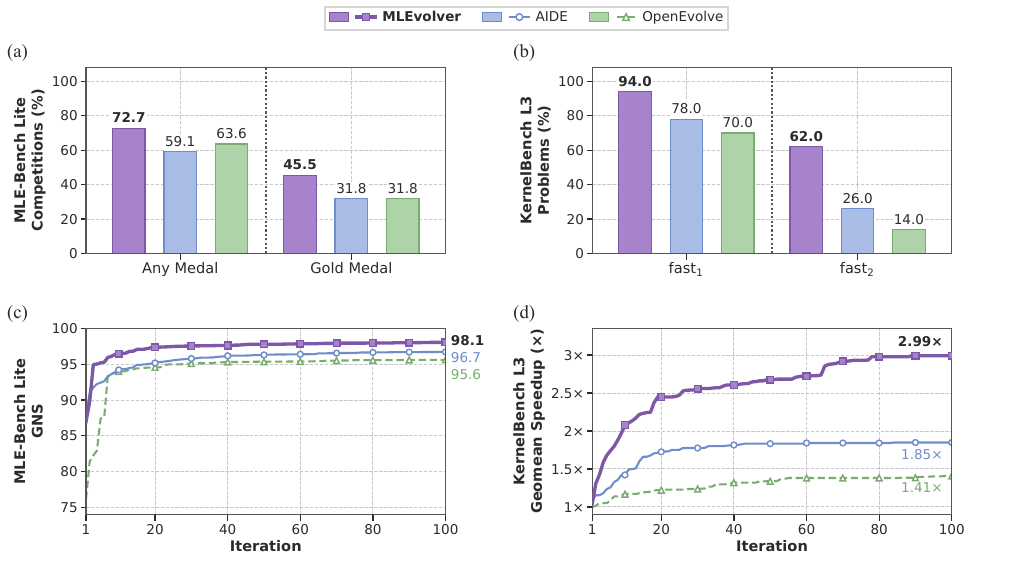}
  \caption{Performance of MLEvolver, AIDE, and OpenEvolve. (a)~Any-medal and gold-medal rates across the 22 MLE-Bench Lite competitions. (b)~$\mathrm{fast}_1$ and $\mathrm{fast}_2$ across the 50 KernelBench Level~3 problems. (c)~Gold-normalized score on MLE-Bench Lite through 100 iterations. (d)~Geometric-mean speedup over PyTorch Eager on KernelBench Level~3 through 100 iterations.}
  \label{fig:mlevolver_results_fig}
\end{figure}

\subsection{Setup}
\label{sec:mlevolver-setup}

\paragraph{Harnesses.}
MLEvolver combines the best-performing mechanism on average in each dimension: folding for context management and deletion for skill management. AIDE~\citep{jiang2025aide} searches a tree of candidate solutions. At each step, it drafts a new solution, refines the best node found so far, or debugs a failing node. OpenEvolve~\citep{sharma2025openevolve}, an open-source implementation of AlphaEvolve~\citep{novikov2025alphaevolve}, maintains a program database organized into islands. At each iteration, it samples a parent program and provides high-performing and ``inspiration'' programs as context. The model edits the parent using search-and-replace diffs.

\paragraph{Benchmarks and budget.}
We evaluate 22 competitions in MLE-Bench Lite, the low-complexity split of MLE-Bench~\citep{chan2025mlebench}, and 50 problems in KernelBench Level~3, which covers full model architectures~\citep{ouyang2025kernelbench}. All three harnesses use \texttt{gpt-6-sol}, a 32K-token context window, and a budget of 100 iterations per run.

\paragraph{Metrics.}
We track progress over iterations using the gold-normalized score (GNS) on MLE-Bench Lite and geometric-mean speedup over PyTorch Eager on KernelBench, as in Section~\ref{sec:experiments}. We also report any-medal and gold-medal rates on MLE-Bench Lite, the percentages of competitions in which the best submission reaches at least the bronze and gold thresholds, respectively~\citep{chan2025mlebench}. On KernelBench, we report $\mathrm{fast}_1$ and $\mathrm{fast}_2$, the percentages of problems with a correct kernel whose speedup over PyTorch Eager exceeds $1\times$ and $2\times$, respectively~\citep{ouyang2025kernelbench}.

\subsection{Results}
\label{sec:mlevolver-results}

\paragraph{MLEvolver outperforms AIDE and OpenEvolve on both benchmarks.}
As illustrated in Figure~\ref{fig:mlevolver_results_fig}a and Figure \ref{fig:mlevolver_results_fig}b, MLEvolver earns medals in 16 of 22 MLE-Bench Lite competitions (72.7\%), a relative gain of 23.1\% over AIDE and 14.3\% over OpenEvolve. It wins 10 gold medals, 42.9\% more than either baseline. On KernelBench Level~3, MLEvolver reaches 94\% on $\mathrm{fast}_1$, a relative gain of 20.5\% over AIDE and 34.3\% over OpenEvolve. At the stricter $\mathrm{fast}_2$ threshold, MLEvolver produces correct kernels with more than $2\times$ speedup over PyTorch Eager on 31 of 50 problems (62\%), more than twice as many as AIDE and more than four times as many as OpenEvolve. These results show that MLEvolver is effective for long-horizon, self-evolving autoresearch on machine learning tasks.

\paragraph{MLEvolver pulls ahead early and extends its lead.} 
Figure~\ref{fig:mlevolver_results_fig}c and Figure \ref{fig:mlevolver_results_fig}d show that MLEvolver reaches 97.4 GNS by iteration 20 on MLE-Bench Lite and a $2.08\times$ geometric-mean speedup by iteration 10 on KernelBench Level~3. On both benchmarks, these early results exceed the performance that AIDE and OpenEvolve reach at iteration 100. On MLE-Bench Lite, MLEvolver maintains its lead and finishes at 98.1 GNS, 1.4\% higher than AIDE and 2.6\% higher than OpenEvolve. On KernelBench, MLEvolver widens its lead in the second half of the run. Between iterations 50 and 100, its geometric-mean speedup increases by 11.9\%, from $2.67\times$ to $2.99\times$, while AIDE and OpenEvolve improve by 0.8\% and 5.2\%, respectively.

\subsection{Discussion}
\paragraph{Folded history supports informed exploration and refinement.}
All three systems can explore different approaches. AIDE~\citep{jiang2025aide} follows a predefined policy for drafting, debugging, and improving solutions, with hyperparameters controlling the number of initial drafts and debugging behavior. OpenEvolve~\citep{sharma2025openevolve} samples parent and inspiration programs from its populations according to configured rules. In MLEvolver, the model uses a catalog of attempt digests to decide whether to refine a promising solution, debug a failure, or propose a new approach when progress stalls. It can also retrieve full records of relevant earlier attempts, including their code, recorded reasoning, scores, and execution output. This combination provides an overview of previously explored directions and access to the details needed to build on earlier successes and failures. The value of retaining this access is reflected in Section~\ref{sec:analysis-folding}, where folding preserves more experimental history within the context window and achieves a higher new-best rate during later iterations.

\paragraph{Distilled skills give MLEvolver direct guidance for later attempts.}
AIDE summarizes earlier valid solutions, and OpenEvolve provides selected programs alongside summaries of recent changes, metrics, and outcomes. Alongside the attempt digests, MLEvolver maintains a separate library of lessons distilled from its experiments. Retrieving these lessons directly reduces the need to reconstruct them from earlier code and results. Offensive skills suggest promising directions to explore, while defensive skills record constraints and failure patterns that can help the model avoid repeating mistakes. Deletion prunes persistently unselected skills, using the model's retrieval behavior as a signal of which lessons remain relevant as the search progresses. The benefits of both retaining and managing these lessons are supported by Section~\ref{sec:analysis-gain-decomposition}, where maintaining the skill library improves performance and deletion provides a further gain.

\paragraph{Solution space shapes scaling.}
The two benchmarks differ in the structure of their solution spaces. On MLE-Bench Lite, the solution space is more discrete, with promising solutions built around distinct choices of model family and feature representation. The harnesses can therefore start with substantially different submissions, contributing to the spread in GNS at iteration 1.
On KernelBench, the solution space is more continuous. The first correct kernels run about as fast as PyTorch Eager on average, and later optimizations yield incremental speedups. These differences favor different search strategies. On MLE-Bench Lite, exploring several distinct candidates increases the chance of finding a strong solution. AIDE explores candidates across branches of a search tree, and OpenEvolve samples from a population of programs. As the runs progress, AIDE's gap to MLEvolver narrows from 3.5\% to 1.4\%, and OpenEvolve's narrows from 15.4\% to 2.6\%. On KernelBench, successive optimizations can build on one another, favoring sequential self-refinement. MLEvolver uses a single refinement chain with context folding and skill deletion, so each new edit can draw on results and lessons from earlier attempts. AIDE refines a node in its search tree, and OpenEvolve edits a sampled parent with guidance from selected programs. In both cases, the model has a more limited view of which optimizations succeeded or failed in earlier attempts. Between iterations 50 and 100, MLEvolver increases its geometric-mean speedup from $2.67\times$ to $2.99\times$, while AIDE and OpenEvolve make smaller gains and finish at $1.85\times$ and $1.41\times$, respectively.

\section{Conclusion}
We present a controlled study of harness design for long-horizon, self-evolving autoresearch agents. Within one agent, we isolate five memory mechanisms and compare each with a recency-truncation baseline on MLE-Bench and KernelBench using \texttt{gpt-oss-120b}, \texttt{gpt-6-luna}, and \texttt{gpt-6-sol}. Three findings
emerge from the experiment. First, folding performs best in all settings by condensing each attempt into a short digest, allowing the agent to retain more experimental history within the context budget and target solutions more effectively. Second, both retaining skills and actively managing them improve performance, with deletion providing the largest average gain among skill-management mechanisms. Third, the benefits of skill management vary with the model backbone's skill selection behavior. Both \texttt{gpt-6-luna} and \texttt{gpt-6-sol} retrieve fewer skills and favor defensive guidance more strongly than \texttt{gpt-oss-120b}, accompanying their smaller gains from skill management. Based on these findings, we combine folding and deletion in MLEvolver, which achieves a 72.7\% medal rate on MLE-Bench Lite and 94\% on $\mathrm{fast}_1$ on KernelBench Level~3, outperforming AIDE and OpenEvolve on both benchmarks. Overall, our study identifies effective choices for preserving experimental history and curating reusable lessons, providing an empirically grounded foundation for long-horizon, self-evolving AI research agents.
\section*{Limitations}

Our study isolates memory mechanisms within linear, sequential refinement loops and leaves parallel scaling unexplored. Parallel search strategies, such as concurrent tree search and population-based evolution, distribute attempts across parallel trajectories and place distinct demands on context budgets and skill selection. Our system comparisons cover structured autoresearch frameworks with guided research loops. Interactive coding agents, such as Claude Code or Codex, are not evaluated. They execute open-ended sequences of tool calls and internal refinement rounds, making it difficult to define a common unit of iteration or enforce comparable evaluation budgets across systems. Finally, each task initializes an independent skill library. Our findings therefore characterize within-task experience accumulation, leaving other evolutionary effects, such as cross-domain skill transfer and multi-task skill distillation, for future work.
\bibliography{references}
\clearpage
\appendix
\section{KernelBench Protocol and Evaluation Details}
\label{sec:kernelbench_details}

We evaluate all agent configurations on 50 tasks from KernelBench \citep{ouyang2025kernelbench}. This section describes the evaluation environment, timing procedure, task subset selection, and reward-hacking safeguards.

\subsection{Evaluation Environment and Timing Protocol}
\label{app:kb-environment}

All evaluations execute on a dedicated NVIDIA GH200 144G HBM3e GPU. The software environment runs Python 3.10.20, PyTorch 2.11.0+cu128, and CUDA 12.8.

\paragraph{Equivalence Verification.}
Following \citet{ouyang2025kernelbench}, functional correctness evaluates numerical output equivalence between the PyTorch reference implementation and the agent's submitted kernel on identical input tensors.

\paragraph{Latency Measurement.} 
The original KernelBench protocol \citep{ouyang2025kernelbench} uses the arithmetic mean execution time across runs to compute speedup, as its benchmark suite exhibits a coefficient of variation ($\sigma/\mu$) below 3\%. In our long-horizon agent setting, execution times do not consistently remain within this variation threshold across repeated compilation and execution passes. We therefore use the median execution time to compute speedup, reducing the influence of transient operating-system overhead and driver allocation spikes.

\subsection{Stratified Problem Selection}
\label{app:kb-sampling}

KernelBench contains 250 problems across three difficulty levels. The full suite includes many similar operations evaluated with different tensor shapes, so uniform sampling could overrepresent some computational patterns. We therefore construct a 50-task subset from 28 strata defined by computational structure. The subset includes at least one task from every stratum and preserves the benchmark's 2:2:1 level distribution: 20 Level-1 tasks testing individual operators, 20 Level-2 tasks testing fused operator chains, and 10 Level-3 tasks testing model architectures. Table~\ref{tab:kb-manifest} lists the selected problems and their stratum assignments.

\paragraph{Sampling Frame and Exclusions.}
Our sampling frame contains 238 of the 250 KernelBench problems: 99 from Level 1, 89 from Level 2, and all 50 from Level 3. We exclude 12 problems for the following reasons:

\begin{itemize}
    \item \textbf{Degenerate problems (3 Level-2 tasks: p23, p80, and p83).}
    The reference implementations produce outputs that are either identically zero or numerically indistinguishable from zero under the benchmark's correctness tolerance. For example, in problem 80, a column-wise maximum reduces the tensor to a single column, after which mean subtraction and GELU produce an all-zero output. A trivial kernel that returns zeros can therefore pass numerical validation without performing the intended computation. We exclude these problems because they do not meaningfully evaluate kernel optimization.

    \item \textbf{Algebraically reducible problems (8 Level-2 tasks: p13, p14, p18, p42, p44, p51, p65, and p98).}
    In these problems, a matrix multiplication or convolution is followed by operations that permit most of the original computation to be removed through algebraic simplification. For example, the reference computation in problem 14, $\sum_j (xW^\top)_j / 2 \cdot s$, can be rewritten as $x\left(\sum_j W_j\right) / 2 \cdot s$. This reformulation replaces a general matrix multiplication (GEMM) with a less expensive general matrix-vector multiplication (GEMV). We exclude these problems because their performance is determined primarily by discovering an algebraic shortcut.

    \item \textbf{Missing reference timing (1 Level-1 task: p95).}
    Problem p95, \texttt{CrossEntropyLoss}, has no valid FP32 reference latency in our evaluation procedure. Because speedup is defined relative to the PyTorch Eager reference latency, we cannot compute a speedup for this problem and therefore exclude it from the sampling frame.
\end{itemize}

\begin{table}[!htbp]
\centering
\scriptsize
\setlength{\tabcolsep}{2.8pt}
\renewcommand{\arraystretch}{0.96}

\begin{tabularx}{\textwidth}{
@{}rr>{\raggedright\arraybackslash}Xl@{\hspace{5pt}}rr>{\raggedright\arraybackslash}Xl@{}
}
\toprule
\textbf{L} & \textbf{ID} & \textbf{Task} & \textbf{Stratum}
&
\textbf{L} & \textbf{ID} & \textbf{Task} & \textbf{Stratum} \\
\midrule

1 & 3
& Batched matrix multiplication
& \texttt{dense-linalg}
&
2 & 8
& Conv3d, max pool, and reduction
& \texttt{l2:conv3d:reduce} \\

1 & 13
& Symmetric matrix multiplication
& \texttt{dense-linalg}
&
2 & 10
& Transposed Conv2d and reduction
& \texttt{l2:conv:reduce} \\

1 & 22
& Tanh
& \texttt{pointwise}
&
2 & 16
& Transposed Conv2d with pointwise operations
& \mcell{\texttt{l2:conv3d:pointwise+}\\
                 \texttt{l2:convT2d:pointwise}} \\

1 & 34
& Instance normalization
& \texttt{norm-library+norm-manual}
&
2 & 20
& Transposed Conv3d with residual operations
& \texttt{l2:convT3d:pointwise} \\

1 & 43
& Three-dimensional max pooling
& \texttt{pool-max+pool-avg}
&
2 & 26
& Transposed Conv3d and HardSwish
& \texttt{l2:convT3d:pointwise} \\

1 & 47
& Sum over one dimension
& \texttt{row-reduce}
&
2 & 30
& GEMM, GroupNorm, and Hardtanh
& \texttt{l2:gemm:norm} \\

1 & 53
& Minimum over one dimension
& \texttt{row-reduce}
&
2 & 31
& Conv2d with pointwise operations
& \texttt{l2:conv2d:pointwise} \\

1 & 57
& Two-dimensional transposed convolution
& \texttt{convT1d+convT2d}
&
2 & 40
& GEMM with scaling and residual addition
& \texttt{l2:gemm:pointwise} \\

1 & 58
& Three-dimensional transposed convolution
& \texttt{convT3d}
&
2 & 45
& GEMM, sigmoid, and LogSumExp
& \texttt{l2:gemm:reduce} \\

1 & 63
& Two-dimensional convolution
& \texttt{conv-direct}
&
2 & 48
& Conv3d with pointwise operations
& \mcell{\texttt{l2:conv3d:pointwise+}\\
                 \texttt{l2:convT2d:pointwise}} \\

1 & 72
& Strided grouped transposed Conv3d
& \texttt{convT3d}
&
2 & 55
& GEMM, max pooling, and reduction
& \texttt{l2:gemm:reduce} \\

1 & 75
& Grouped dilated transposed Conv2d
& \texttt{convT1d+convT2d}
&
2 & 57
& Conv2d, ReLU, and HardSwish
& \texttt{l2:conv2d:pointwise} \\

1 & 81
& Dilated transposed Conv2d
& \texttt{convT1d+convT2d}
&
2 & 68
& GEMM with minimum and subtraction
& \texttt{l2:gemm:pointwise} \\

1 & 82
& Two-dimensional depthwise convolution
& \texttt{conv-direct}
&
2 & 73
& Conv2d and batch normalization
& \texttt{l2:conv:norm} \\

1 & 83
& Asymmetric depthwise convolution
& \texttt{conv-direct}
&
2 & 77
& Transposed Conv3d and batch normalization
& \texttt{l2:convT3d:norm} \\

1 & 92
& Exclusive cumulative sum
& \texttt{scan}
&
2 & 78
& Transposed Conv3d and reductions
& \texttt{l2:conv:reduce} \\

1 & 93
& Masked cumulative sum
& \texttt{scan}
&
2 & 79
& Conv3d and instance normalization
& \texttt{l2:conv:norm} \\

1 & 97
& Scaled dot-product attention
& \texttt{attention-sdpa}
&
2 & 81
& GEMM with pointwise operations
& \texttt{l2:gemm:pointwise} \\

1 & 99
& Triplet-margin loss
& \texttt{loss-reduce}
&
2 & 88
& GEMM, GroupNorm, and Swish
& \texttt{l2:gemm:norm} \\

1 & 100
& Hinge loss
& \texttt{loss-reduce}
&
2 & 94
& GEMM, activations, and GroupNorm
& \texttt{l2:gemm:norm} \\

\midrule

3 & 2
& Shallow wide MLP
& \texttt{mlp-wide-step+mlp}
&
3 & 35
& Six-layer LSTM
& \mcell{\texttt{recurrent-loop+}\\
                 \texttt{recurrent-cudnn}} \\

3 & 7
& GoogLeNet
& \texttt{cnn-full}
&
3 & 44
& MiniGPT block
& \mcell{\texttt{attention-block+}\\
                 \texttt{transformer-full}}\\

3 & 9
& ResNet-18
& \texttt{cnn-full}
&
3 & 47
& NetVLAD without ghost clusters
& \texttt{pooling-agg+ssm} \\

3 & 12
& VGG-19
& \texttt{cnn-small+cnn-classic}
&
 & & & \\

3 & 25
& ShuffleNet unit
& \texttt{cnn-block}
&
 & & & \\

3 & 27
& RegNet
& \texttt{cnn-full}
&
 & & & \\

3 & 32
& Convolutional Vision Transformer
& 
\mcell{\texttt{attention-block+}\\
                 \texttt{transformer-full}}
&
 & & & \\

\bottomrule
\end{tabularx}

\caption{The 50-task KernelBench evaluation subset and its sampling strata.
The subset contains 20 Level-1, 20 Level-2, and 10 Level-3 tasks and covers
all 28 strata. Task labels are shortened for readability; IDs refer to the
original KernelBench problem identifiers.}
\label{tab:kb-manifest}
\end{table}

\paragraph{Stratification and Sample Allocation.}
We group the 238 eligible tasks into 28 strata according to their computational structure. The strata distinguish, for example, reductions over different tensor dimensions, elementwise operations following matrix multiplications or convolutions, and sequential or recurrent computations.

We determine the initial number of tasks assigned to each stratum using a regularized form of Neyman allocation \citep{neyman1934}:
\begin{equation}
    n_h
    =
    n
    \frac{N_h\hat{\sigma}_h}
         {\sum_{k=1}^{H}N_k\hat{\sigma}_k},
\end{equation}
where $H=28$ is the number of strata, $n=50$ is the total sample size, $N_h$ is the number of eligible tasks in stratum $h$, and $n_h$ is the number sampled from that stratum. The quantity $\hat{\sigma}_h$ estimates the within-stratum standard deviation of $\log(\mathrm{speedup})$ across pilot replicates. We regularize this estimate toward an analytic optimization-headroom score ranging from 0 to 3. We also require every stratum to contain at least one sampled task.

The level budgets are fixed at 20, 20, and 10 tasks, matching the full benchmark's 2:2:1 distribution. Sampling weights account for this difference when estimating performance over the eligible frame.

\paragraph{Subset Selection and Runtime Matching.} 
Using the stratum allocations, we generate 3,000 candidate subsets by sampling uniformly without replacement within each stratum, with seed \texttt{20260919}. We select the candidate that best matches the 238-task sampling frame in log reference runtime, using an objective that combines the Kolmogorov--Smirnov (KS) distance, the difference in mean log runtime, and deviation from the target level distribution. Reference runtime is the median latency of the PyTorch eager implementation on the evaluation hardware and is independent of agent performance.

\subsection{Anti-Reward-Hacking Safeguards}
\label{app:kb-antihack}

Kernel-generation agents can discover shortcuts that increase measured
speedup without implementing the intended computation
\citep{baronio2026kevin,dai2026cudaagentlargescaleagentic}. We combine static
code inspection, runtime isolation, and prompt-based guidance to detect or
discourage these strategies.

\paragraph{Reference Output Isolation.}
Prior work has shown that candidate kernels may exploit outputs left in GPU
memory by the preceding reference execution \citep{baronio2026kevin}. In
particular, a candidate that allocates an uninitialized tensor may receive
memory containing values from the reference output and pass numerical
verification without writing every output element. We prevent this shortcut
by retaining the reference output until numerical verification is complete,
so its device memory cannot be reused during candidate execution.

\paragraph{Static Code Inspection.}
Before compilation, we inspect each candidate implementation using Python's
\texttt{ast} module. The validation gate rejects candidates that:

\begin{enumerate}
    \item do not define explicit CUDA source containing at least one
    \texttt{\_\_global\_\_} kernel;
    \item delegate all reference operations to high-level frameworks or
    compilation interfaces instead of implementing a complete operation in a
    custom kernel;
    \item patch timing functions, reference modules, or evaluation code;
    \item modify input tensors or tensor references in place; or
    \item spawn background threads or processes, or defer timed work to an
    execution stream that is not synchronized with the primary stream.
\end{enumerate}

A candidate that fails any of these checks is marked invalid and can never become the task's best attempt; speedup is computed from each task's best valid attempt.

\paragraph{Numerical State Isolation.}
Candidates also attempt to alter the evaluation by changing process-wide
numerical settings, such as PyTorch matrix-multiplication precision or global
TF32 flags. The harness records the relevant numerical settings and restores
the required evaluation configuration before reference timing. Candidates
that modify prohibited precision or backend settings are rejected. This
prevents a candidate from obtaining an apparent speedup by reducing numerical
precision or by slowing the reference implementation.

\paragraph{Prompt-Based Integrity Guidance.}
\label{app:kb-prompt}
The runtime safeguards are complemented by prompt instructions that discourage invalid optimization strategies before code is generated or reusable knowledge is recorded. These instructions guide the agent toward genuine kernel optimization but are not treated as enforcement mechanisms.
\par
\begin{tcolorbox}[
    enhanced,
    breakable,
    title=\textbf{KernelBench Integrity Rules},
    colback=blue!3,
    colframe=blue!55!black,
    colbacktitle=blue!55!black,
    coltitle=white,
    fonttitle=\small\bfseries,
    boxrule=0.7pt,
    arc=2mm,
    left=1.5mm,
    right=1.5mm,
    top=1mm,
    bottom=1mm,
    before skip=4pt,
    after skip=4pt
]
\small
\begin{enumerate}
    \setlength{\itemsep}{2pt}
    \setlength{\parskip}{0pt}
    \setlength{\parsep}{0pt}

    \item \textbf{Compute the intended operation.}
    At least one custom CUDA kernel must implement a complete operation or
    fused group from the reference model. Dummy kernels, identity operations,
    and staging copies do not qualify.

    \item \textbf{Preserve correctness beyond observed inputs.}
    Implementations may specialize to the declared shapes, dtypes, constructor
    arguments, and target GPU, but not to particular input values, weights,
    memory addresses, call counts, or random seeds.

    \item \textbf{Use only exact simplifications.}
    Algebraic rewrites are allowed when they preserve the reference computation
    for every valid input and weight. Approximations and shortcuts that hold
    only for observed test inputs are not allowed.

    \item \textbf{Preserve the required numerical precision.}
    Implementations may not introduce lower-precision arithmetic, approximate
    mathematical operations, or prohibited changes to process-wide precision
    settings.

    \item \textbf{Do not manipulate evaluation.}
    Implementations may not modify inputs, reference buffers, timers, backend
    settings, execution streams, threads, or processes to influence correctness
    or latency measurements.
\end{enumerate}
\end{tcolorbox}

\FloatBarrier
\section{MLE-Bench Protocol and Evaluation Details}
\label{mle-modification}

\begin{table}[!htbp]
\centering
\footnotesize
\setlength{\tabcolsep}{4pt}
\begin{tabularx}{\columnwidth}{@{}Xrr@{}}
\toprule
\textbf{Competition} & \textbf{Train} & \textbf{Test} \\
\midrule
\multicolumn{3}{@{}l}{\textit{Low complexity}} \\
denoising-dirty-documents & 115 & 29 \\
detecting-insults-in-social-commentary & 3,947 & 2,647 \\
mlsp-2013-birds & 258 & 64 \\
random-acts-of-pizza & 2,878 & 1,162 \\
tabular-playground-series-may-2022 & 800,000 & 100,000 \\
the-icml-2013-whale-challenge-right-whale-redux & 22,692 & 25,149 \\
\midrule
\multicolumn{3}{@{}l}{\textit{Medium complexity}} \\
lmsys-chatbot-arena & 51,729 & 5,748 \\
statoil-iceberg-classifier-challenge & 1,283 & 321 \\
tweet-sentiment-extraction & 24,732 & 2,749 \\
us-patent-phrase-to-phrase-matching & 32,825 & 3,648 \\
\bottomrule
\end{tabularx}
\caption{The ten analyzed competitions, by MLE-Bench identifier and
complexity label. Elsewhere we shorten the identifiers (e.g., whale-redux).
Train and test sizes count the instances in MLE-Bench's splits: images
(denoising-dirty-documents, statoil-iceberg), recordings (mlsp-2013-birds),
audio clips (whale-redux), table rows (tabular-may-2022), and text examples
for the other competitions.}
\label{tab:mle-competitions}
\end{table}

\subsection{Competition Composition}
\label{app:mle-competitions}
We evaluated ten MLE-Bench competitions \citep{chan2025mlebench} and analyzed all of them.

\paragraph{Subset selection.}
Before final benchmarking, we selected six low-complexity and four medium-complexity MLE-Bench competitions spanning tabular, text, audio, and vision tasks. Selection was based on data diversity and did not use configuration rankings.

\paragraph{Data.}
We used MLE-Bench’s own preparation scripts and splits, and no data were excluded or modified.

\begin{table}[!htbp]
\centering
\footnotesize
\begin{tabular}{@{}llrrrrrrc@{}}
\toprule
Short name & Metric & $s_0$ & $s^\ast$ & $s_{\mathrm{gold}}$ & Teams & Gold rank & $C^{\mathrm{gold}}$ & $w$ \\
\midrule
denoising-dirty-documents & RMSE $\downarrow$     & 0.26461 & 0 & 0.01794 & 162   & 10 & 0.9322 & 1 \\
detecting-insults        & AUC $\uparrow$         & 0.5     & 1 & 0.83321 & 50    & 5  & 0.6664 & 1 \\
mlsp-2013-birds          & AUC $\uparrow$         & 0.5     & 1 & 0.93527 & 81    & 8  & 0.8705 & 1 \\
random-acts-of-pizza     & AUC $\uparrow$         & 0.5     & 1 & 0.97908 & 462   & 10 & 0.9582 & 1 \\
tabular-may-2022         & AUC $\uparrow$         & 0.5     & 1 & 0.99823 & 1,152 & 12 & 0.9965 & 1 \\
whale-redux              & AUC $\uparrow$         & 0.5     & 1 & 0.98961 & 129   & 10 & 0.9792 & 1 \\
lmsys-chatbot-arena      & log loss $\downarrow$  & 1.096   & 0 & 0.98392 & 1,688 & 13 & 0.1023 & 2 \\
statoil-iceberg          & log loss $\downarrow$  & 0.69314 & 0 & 0.11374 & 3,330 & 16 & 0.8359 & 2 \\
tweet-sentiment          & Jaccard $\uparrow$     & 0.5932  & 1 & 0.72689 & 2,225 & 14 & 0.3286 & 2 \\
us-patent                & Pearson $r$ $\uparrow$ & 0       & 1 & 0.87000 & 1,889 & 13 & 0.8700 & 2 \\
\bottomrule
\end{tabular}
\caption{Evaluation and gold anchoring. $\uparrow$: maximized;
$\downarrow$: minimized. $s_0$: do-nothing score; $s^\ast$: perfect score;
$s_{\mathrm{gold}}$: gold-medal threshold, at the given rank among the given
number of teams on the private leaderboard. $C^{\mathrm{gold}}$: normalized
score of the gold threshold. $w$: weight in GNS.}
\label{tab:mle-evaluation}
\end{table}

\subsection{Competition-Specific Evaluation}

Table~\ref{tab:mle-evaluation} gives the metric and the direction, the do-nothing score $s_0$, the perfect score $s^\ast$, and the gold medal threshold $s_{\mathrm{gold}}$ of each competition. 

\paragraph{Do-nothing score.}
For AUC, $s_0 = 0.5$, since it is a random ranking. For Pearson correlation, $s_0 = 0$, since the metric is already zero for an uninformative prediction. For the other metrics we measure $s_0$ by grading the do-nothing submission with the competition's own grader: submitting each tweet unchanged (Jaccard), predicting the class frequencies (log loss, so $s_0$ is the entropy of the label distribution), and predicting the best constant, the mean pixel value (RMSE).

\paragraph{Gold threshold.}
$s_{\mathrm{gold}}$ is the score at the gold rank on the competition's private leaderboard snapshot distributed with MLE-Bench, under Kaggle's medal rule: the top 10\% of teams when fewer than 100 teams entered, the top 10 teams for 100 to 249 teams, and the top 10 teams plus 0.2\% of all teams for 250 or more. We take $s_{\mathrm{gold}}$ from the leaderboard that MLE-Bench provides for each competition \citep{chan2025mlebench}, a snapshot of the competition's final Kaggle private leaderboard.

\subsection{Raw Score Normalization.}
\label{app:mle-normalization}

The competitions use different metrics with different directions, so we convert a run's best raw score $s_i$ on competition $i$ into a gold-normalized score:
\begin{equation}
  \mathrm{GNS}_i = 100 \cdot \frac{s_i - s_i^{0}}{s_i^{\mathrm{gold}} - s_i^{0}},
\end{equation}
where $s_i^{0}$ is the score of a do-nothing submission and $s_i^{\mathrm{gold}}$ the gold-medal threshold (Table~\ref{tab:mle-evaluation}). For minimized metrics (RMSE and log loss), the numerator and the denominator change sign together, so a do-nothing submission scores 0 and a gold-level one 100 regardless of the metric’s direction. This is the form of human-normalized score in reinforcement learning \citep{mnih2015human}.

\paragraph{Invalid or missing submissions.}
An attempt that crashes, exceeds its limits, or produces no valid submission receives no score. It does not count toward the run's best, but it is considered as using up one iteration. A run's score on a competition is that of its best valid attempt. 

\subsection{Gold Anchoring}
\label{app:mle-gold}

We anchor at the gold medal threshold because the distance from gold to perfect varies widely across competitions. Measured against gold, every competition has the same reference point, 100, set by the best human teams on MLE-Bench's leaderboard \citep{chan2025mlebench}.

\paragraph{Scores above gold are retained.}
We do not clip $\mathrm{GNS}_i$ at 100. If the competition beats the gold threshold, the score may be over 100. 

\subsection{Complexity Weights}
\label{app:mle-weights}

For each configuration, we take $\mathrm{GNS}_i$ over its runs, and aggregate over the ten analyzed competitions with a weighted mean:

\begin{equation}
  \mathrm{GNS} = \frac{\sum_i w_i\, \mathrm{GNS}_i}{\sum_i w_i},
\end{equation}
with $w_i = 1$ for low-complexity and $w_i = 2$ for medium-complexity competitions. The categories come directly from MLE-Bench. MLE-Bench labels a competition low complexity if an experienced machine learning engineer could produce a sensible solution in under two hours, excluding training time. It labels a competition medium complexity if it would take two to ten hours \citep{chan2025mlebench}. Based on that, we weight medium complexity competitions twice because they demand more engineering. 

\subsection{Execution Protocol}
\label{app:mle-execution}

Each run starts from a deliberately chance-level seed solution, the same for every configuration and backbone, which is evaluated as attempt 0. The agent then runs 100 iterations, each consisting of one coder call and one execution of the resulting solution.

Before grading, \texttt{submission.csv} must exist and match the sample submission's columns and row count. MLE-Bench's grader then scores it, rejecting malformed files  \citep{chan2025mlebench}. An attempt that crashes, times out, or produces an invalid submission receives no score, and the agent sees the error output in its next prompt.

\newpage
\FloatBarrier
\section{Evolution Dynamics on Each Design}
\begin{figure*}[htp]
  \centering
  \includegraphics[width=0.97\textwidth]{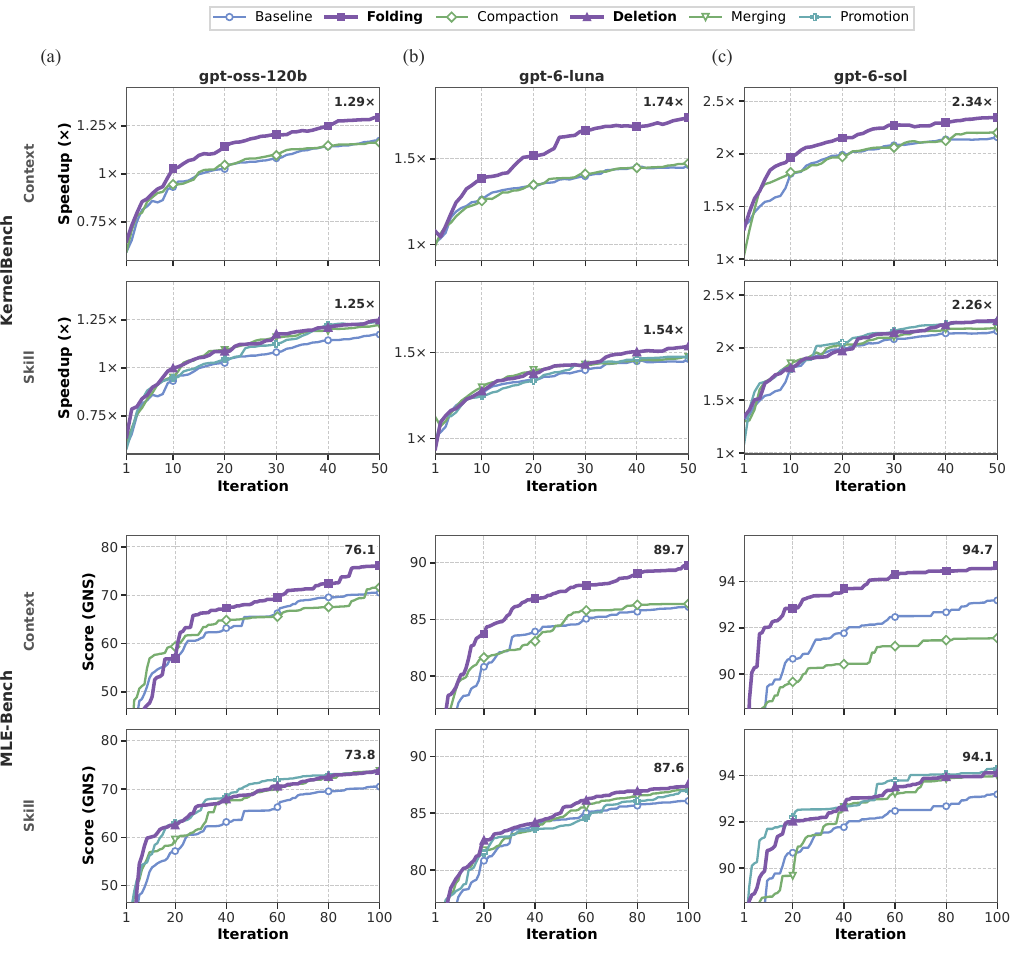}
  \caption{Best result so far over iterations for each design.
  Columns: (a) gpt-oss-120b, (b) gpt-6-luna, (c) gpt-6-sol.
  Rows: context management (baseline, folding, compaction) and
  skill management (baseline, deletion, merging, promotion), on
  KernelBench (top two rows) and MLE-Bench (bottom two rows).
  KernelBench: geometric mean, over 50 problems, of the speedup
  over PyTorch eager of the best kernel within the first $N$
  attempts. MLE-Bench: gold-normalized score (GNS; 0 = chance,
  100 = gold) of the best solution so far, averaged over 10
  competitions.}
  \label{fig:evolution}
\end{figure*}

Figure~\ref{fig:evolution} illustrates the cumulative best performance over iterations for each design across three backbones (\texttt{gpt-oss-120b}, \texttt{gpt-6-luna}, and \texttt{gpt-6-sol}).

\FloatBarrier
\newpage
\section{Effect of Context Window Size}
\label{app:context-window-size}

\begin{figure}[!htp]
    \centering
    \includegraphics[width=\linewidth]{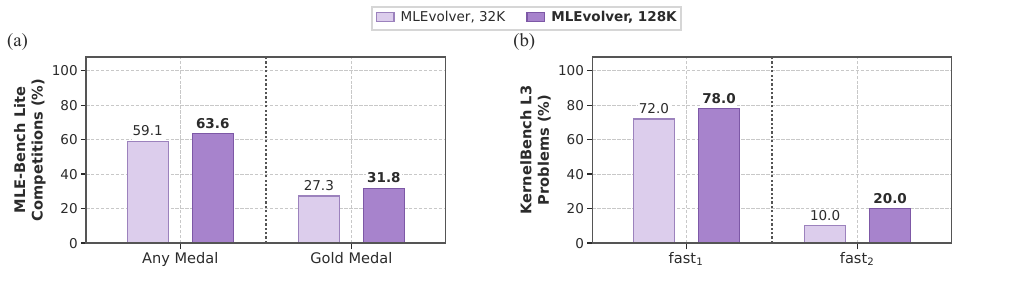}
    \caption{MLEvolver performance with 32K- and 128K-token context windows. (a)~Any-medal and gold-medal rates on MLE-Bench Lite. (b)~$\mathrm{fast}_1$ and $\mathrm{fast}_2$ on KernelBench Level~3.}
    \label{fig:context-window-size}
\end{figure}

Long-context models give autoresearch agents more room for experiment history and retrieved skills. To test whether MLEvolver benefits from this additional capacity, we evaluate it with 32K- and 128K-token context windows under gpt-6-luna and 100 iterations. Figure~\ref{fig:context-window-size} shows that on MLE-Bench Lite, the any-medal rate rises from 59.1\% to 63.6\% and the gold-medal rate from 27.3\% to 31.8\%. On KernelBench Level~3, $\mathrm{fast}_1$ rises from 72.0\% to 78.0\%, and $\mathrm{fast}_2$ doubles from 10.0\% to 20.0\%. With context folding and skill deletion in place, MLEvolver benefits from the larger window on both benchmarks.

\FloatBarrier
\section{Behavior Data on Each Design}

\subsection{Offensive and Defensive Skills}
\label{app:offense-defense}
To describe the skill each backbone chooses, we label all of them as offensive or defensive by their main purpose. An offensive skill aims to raise the score of a working solution: a modeling technique, feature, ensemble, or tuning step on MLE-Bench; a kernel technique such as fusion, tiling, shared-memory reuse, or vectorized access on KernelBench.

A defensive skill aims to prevent failure or loss. Its main responsibility is to avoid crashes, version errors, time or memory overruns, guard against leakage and overfit, or against incorrect results, imprecise numerics, and timing noise on KernelBench.

\paragraph{Rules.} A skill is read as its title, description, and applicability trigger. The title is lower-cased and its qualifiers are removed. Seven fixed rules are tried in order, and the first that matches decides, so no model is involved. Because the title states the skill's main action, rules 1--6 read only the title:

\begin{enumerate}
    \item \textit{Defensive verb} (defensive): the title starts with a protective verb, such as preserve, protect, keep, validate, guard, gate, avoid, prevent, bound, correct, fix, detect, or diagnose
    \item \textit{Caution} (defensive): a warning word or phrase appears anywhere in the title, such as ``not'', ``unproven'', ``redundant'', ``marginal'', ``requires'', ``fails'', ``loses to'', ``regresses'', or ``diminishing''.
    \item \textit{Plateau change} (offensive): the title contains a plateau word (plateau, stagnation, stall) and a change word (diversify, change, pivot, switch, escape, escalate, alternative)
    \item \textit{Offensive head} (offensive): the head is a score-raising verb, such as add, use, combine, blend, stack, encode, augment, tune, calibrate, switch, or escalate, or the title reads ``replace X with Y''.
    \item \textit{Defensive words} (defensive): a protection or a failure to prevent appears anywhere in the title, such as fallback, incumbent, checkpoint, budget, timeout, error, crash, leakage, or overfitting.
    \item \textit{Offensive words} (offensive): a technique or a positive result appears anywhere in the title, such as ensemble, augmentation, embedding, tuning, transformer, outperform, or improve.
    \item \textit{Body}: otherwise, the description and trigger decide: offensive if their offensive words outnumber their defensive and caution words together, and defensive otherwise.
\end{enumerate}

\paragraph{Results.}  Table~\ref{tab:offense-defense} reports
the offensive share of the skills that reached the coder,
pooled over all configurations across MLE-Bench and KernelBench.
\begin{table}[!htbp]
\centering
\small
\begin{tabular}{@{}lcc@{}}
\toprule
 & MLE-Bench & KernelBench \\
\midrule
\texttt{gpt-oss-120b} & \textbf{34.8} & \textbf{71.6} \\
\texttt{gpt-6-luna}   & 24.4 & 36.2 \\
\texttt{gpt-6-sol}   & 14.6 & 36.1 \\
\bottomrule
\end{tabular}
\caption{Offensive share (\%) of the skills picked into the
coder's prompt, over all configurations}
\label{tab:offense-defense}
\end{table}

\subsection{Library Growth Trajectory}

\begin{figure}[!htbp]
  \centering
  \includegraphics[width=\linewidth,height=.40\textheight,keepaspectratio]{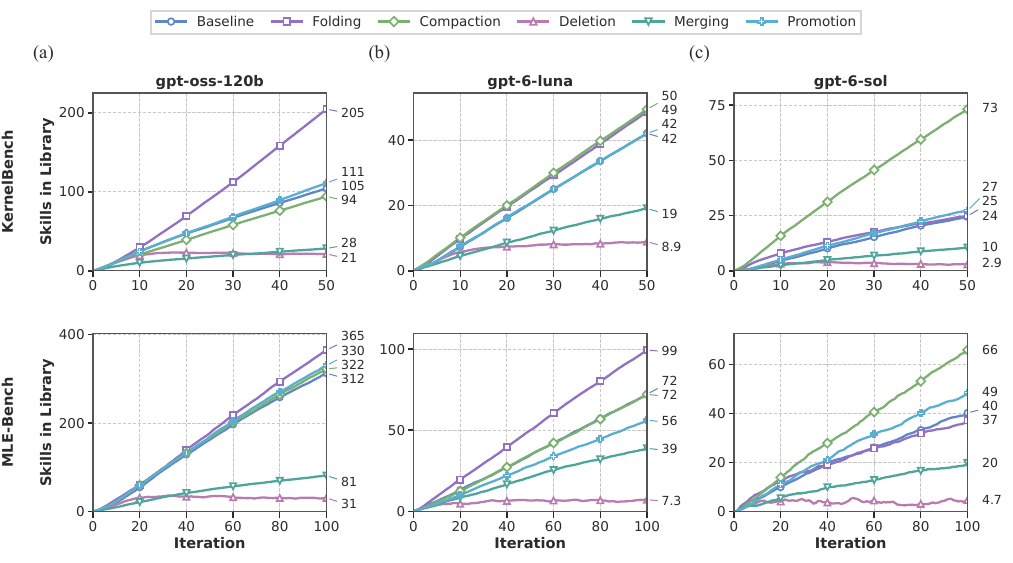}
  \caption{Skills in the library over a run, per competition
  (MLE-Bench) or problem (KernelBench). Column (a): gpt-oss-120b, (b): gpt-6-luna, (c): gpt-6-sol. Without deletion or
  merging, the library grows throughout the run. Deletion
  keeps it small, and merging slows its growth. Promotion's count includes its standing
  rules.}
  \label{fig:skill-library}
\end{figure}

Figure \ref{fig:skill-library} tracks the live skill library through a run: skills written minus those deleted or absorbed by a merge, per competition or problem. Promotion's count includes the skills it has promoted to standing rules.

\FloatBarrier
\subsection{Case Studies of Skill Merging and Promotion}
\label{app:skill-management-cases}
We illustrate merging and promotion with cases on both benchmarks. The boxes summarize the recorded skills and rules. Red marks details weakened or omitted during merging and conditions that cease to hold later in a run.

\newtcolorbox{skillcasebox}[1]{
  enhanced, title={#1},
  colback=white, colframe=promptblueframe, colbacktitle=promptbluetitle,
  coltitle=white, fonttitle=\small\bfseries, fontupper=\small,
  boxrule=0.6pt, arc=1.5mm, left=2mm, right=2mm, top=2mm, bottom=2mm,
  before skip=6pt, after skip=6pt
}

\subsubsection{Merging Can Remove Task-Specific Details}
\label{app:skill-merge-case}

\begin{skillcasebox}{KernelBench Level~1, Problem~3 --- Dense Batched Matrix Multiplication}
\textbf{Before merging}\par\smallskip
\textbf{Skill 9.} Consider a tensor-core implementation only if it preserves \textcolor{loss}{the required FP32 semantics}. \textcolor{loss}{Do not substitute TF32, FP16, or BF16 when FP32 is required. No tensor-core implementation was tested.}\par
\textbf{Skill 10.} Address the scalar-FMA bottleneck while \textcolor{loss}{preserving FP32 semantics}. Retain boundary masks unless divisibility is guaranteed.
\tcblower
\textbf{After merging}\par\smallskip
Use a tensor-core implementation that meets the required \textcolor{loss}{\texttt{PRECISION}} contract. Use lower precision only when the contract permits it. Retain masks unless divisibility is guaranteed.
\end{skillcasebox}

At iteration~10 of this run, merging replaces the explicit FP32 requirement with an unfilled placeholder and omits that the tensor-core approach was untested.

\begin{skillcasebox}{MLE-Bench --- Denoising Dirty Documents}
\textbf{Before merging (source skill 13)}\par\smallskip
When editing nested \textcolor{loss}{\texttt{extend} or \texttt{append} calls containing NumPy expressions}, check for a missing closing parenthesis. In this case, \textcolor{loss}{the nested \texttt{wide\_maps.extend(...)} expression needs one more closing parenthesis}.
\tcblower
\textbf{After merging (skill 11)}\par\smallskip
Balance delimiters from the innermost expression outward and run a syntax check before starting costly processing.
\end{skillcasebox}

In this run, skill~11 absorbed skills~13--15 in three successive merges. The final merged skill retains the general syntax advice but omits the specific expression recorded in skill~13. The same missing-parenthesis error recurs at iteration~79. At iteration~80, the debugging attempt retrieves only the merged skill~11, leaves the faulty expression unchanged, and fails with the same syntax error.

\subsubsection{Promoted Rules Can Outlast Their Applicability}
\label{app:skill-promotion-case}

\begin{skillcasebox}{KernelBench Level~2, Problem~57 --- Convolution with ReLU and HardSwish}
\textbf{Promoted rule}\par\smallskip
When \textcolor{loss}{fusing ReLU and HardSwish in one hand-written CUDA kernel}, preserve FP32 evaluation order and remove only the lower clamp made redundant by ReLU.
\tcblower
\textbf{Later in the run}\par\smallskip
By iteration~20, cuDNN handles ReLU and the separate CUDA kernel computes HardSwish only.
At iterations~30 and~50, the incumbent again fuses both activations in a hand-written convolution.
\end{skillcasebox}

The rule is promoted at iteration~10, when a single CUDA kernel applies both ReLU and HardSwish. Yet, it remains in the prompt as the implementation changes. Its fusion condition no longer holds at iteration~20 but becomes relevant again at iterations~30 and~50. This case illustrates that a promoted rule's applicability can change during a run, while its presence in the prompt remains fixed. \\

\begin{skillcasebox}{MLE-Bench --- Tabular Playground Series (May 2022)}
\textbf{Promoted rule}\par\smallskip
When refining \textcolor{loss}{a LightGBM classifier} on mixed numeric, discrete, and structured-string features, add compact feature summaries and average predictions across seeds.
\tcblower
\textbf{Later in the run}\par\smallskip
LightGBM plateaus at 0.986963 ROC AUC. At iteration~30, the agent switches to a PyTorch network and reaches 0.988587. The network produces all 14 new bests from that point onward and reaches 0.994890.
\end{skillcasebox}

The rule is promoted at iteration~15 to guide LightGBM refinement. It remains active after a neural network becomes the best solution, even though its LightGBM-specific condition no longer holds.

\FloatBarrier
\makeatletter
\setlength{\@fptop}{0pt}
\setlength{\@fpsep}{12pt}
\setlength{\@fpbot}{0pt plus 1fil}
\makeatother
\section{Prompt Templates}
\label{app:prompt-templates}
Figure \ref{fig:prompt-action-selector} to Figure \ref{fig:prompt-promotion} demonstrate the system prompts that define the behavior of the evolving agent across the six configurations. The configurations share the same prompt for action selection, skill retrieval, code implementation, and skill curation, while context folding, compaction, merging, and promotion introduce additional prompts.

\begin{figure}[!htbp]
\centering
\begin{promptbox}{System Prompt: Action Selector}
You are an action-selection engine for an iterative coding agent. L0 is the prior attempt history for THIS problem only. You do NOT select L1 skills in this step. You do NOT write code; you will not be shown any. Choose exactly one next action from the attempt outcomes below.

Return ONLY a JSON object:
{"action": "propose_new|debug_current|refine_current", "rationale": "brief justification"}

Pay special attention to the LAST 3 attempts in full detail when deciding.
\end{promptbox}
\caption{System prompt for the action selection.}
\label{fig:prompt-action-selector}
\end{figure}

\begin{figure}[!htbp]
\centering
\begin{promptbox}{System Prompt: Skill Picker}
You are an L1 skill retrieval engine.
L1 skills are reusable lessons distilled from earlier attempts. Each carries an applicability trigger describing when it is relevant. You are given the attempt history for this problem, the action chosen for this round, and the current symptoms. You do NOT write code; you will not be shown any. Compare the CURRENT symptoms (the latest attempt outcome and terminal output) against each candidate's title, description and applicability trigger, and prefer skills that fit the selected action. Do not re-select a skill whose advice the latest attempts already followed without success.

Each catalog line begins with `- id=<ID>`. Use those IDs verbatim: do not add an "id" prefix and do not renumber them.
Return ONLY a JSON object: {"selected_entry_ids": ["<ID>", "<ID>"]}, or {"selected_entry_ids": []} to select nothing.
No other text or markdown.
\end{promptbox}
\caption{System prompt for skill retrieval.}
\label{fig:prompt-skill-selection}
\end{figure}

\begin{figure}[!htbp]
\centering
\begin{promptbox}{System Prompt: Coder}
You are an expert engineer implementing a solution for the current benchmark task.
The action for this round is already chosen in the user message. Follow the action-specific guidelines section there.

Memory:
- L0: Short-term log for THIS problem (recent attempts in full, older attempts as summaries; use unfolded sections when present).
- L1: Reusable skills distilled from earlier attempts, retrieved for this round (full content below). Follow selected L1 skills as mandatory guidance.
- Iteration budget: see Iteration context in the user message.

Output contract:
1. Return exactly ONE fenced ```python block. No prose outside the fence.
2. Optional: brief diagnosis/hypothesis as # comments at the top inside the fence only.
3. Do not use XML-style tags (<diagnosis>, <action>, <reasoning>, etc.) anywhere in your reply.
[BENCHMARK RULES]
\end{promptbox}
\caption{System prompt for the code implementation.}
\label{fig:prompt-coder}
\end{figure}

\begin{figure}[!htbp]
\centering
\begin{promptbox}{System Prompt: Skill Writer}
You are an Expert Machine Learning Researcher reviewing experiment logs.

Your task is to extract high-density, generalizable insights by performing CONTRASTIVE ANALYSIS on the L0 logs. Do not just summarize; abstract the specific task details into generalizable CUDA/ML engineering principles and insights.

Output EXACTLY this structure:
Title: <a concise, descriptive title for this skill (max 20 tokens)>
Description: <one concise sentence (max 200 tokens) summarizing the details section on the same line as the "Description" heading>
Applicability Trigger: <When should the agent retrieve this memory? Keep it concise and on the same line as the heading.>
Details:
- **The Root Cause:** <Why did the previous attempts fail? If they succeeded, state why and identify the architectural choice that made them work.>
- **The Pivot (Contrast):** <What exact change moved the solution from failure to success or from a lower metric to a better metric?>
- **Generalizable Rule:** <Abstract the pivot into reusable rules. Use symbolic variables rather than task-specific constants.>
- **Anti-Pattern to Avoid:** <Specific patterns that future agents should avoid based on this log.>

Rules:
1. Each part in the Details section can have 1 to 3 bullet points.
2. Do not include raw code blocks.
3. Be technically precise and concise.
4. Avoid generic statements.
\end{promptbox}
\caption{System prompt for skill distillation.}
\label{fig:prompt-skill-writer}
\end{figure}

\begin{figure}[!htbp]
\centering
\begin{promptbox}{System Prompt: Attempt Retrieval (Folding)}
You are choosing which of your earlier attempts to re-read in full.

Each attempt is shown in the catalog as ONE line: the distilled outcome of that attempt. Unfolding an attempt adds its complete record to the next prompt:
  - Action: which of propose_new / debug_current / refine_current it was
  - Progress: whether it set a new best, judged against every attempt before it
  - Skills: the L1 skill ids that were in play for it
  - Reasoning: your own reasoning at the time, in full
  - Code: the complete candidate source, untruncated
  - Terminal: the metrics line plus the raw compiler / runtime output, untruncated
Some fields are absent on some attempts.

The catalog tells you WHAT happened and the unfolded record tells you WHY. Unfold when you need the source or exact error text, not to re-read a conclusion you already have.
Respond with ONLY a JSON object: {"unfold_round_ids": ["round_1"]} or {"unfold_round_ids": []}
Do NOT output Python code or markdown fences.
\end{promptbox}
\caption{System prompt for attempt retrieval in Folding.}
\label{fig:prompt-folding-preflight}
\end{figure}

\begin{figure}[!htbp]
\centering
\begin{promptbox}{System Prompt: Attempt Digest Summarizer (Folding)}
You compress ONE coding attempt from an evolving agent into a short memory line for later prompts.

Focus on: action taken, outcome, root error or bottleneck, and what changed versus prior context.
Do NOT extract reusable L1 skills. Do NOT output JSON or markdown fences.
Return 2-4 sentences of plain text only.
\end{promptbox}
\caption{System prompt for attempt summarization in folding.}
\label{fig:prompt-folding-summarizer}
\end{figure}

\begin{figure}[!htbp]
\centering
\begin{promptbox}{System Prompt: Context Compaction}
You create anchored compaction summaries so another coding agent can continue the work.
Do not call tools. Do not invent scores, files, or outcomes. Preserve exact paths, symbols, error strings, and metric values when known. Do not mention the summary process or that context was compacted.
\end{promptbox}
\caption{System prompt for context compaction.}
\label{fig:prompt-compaction}
\end{figure}

\begin{figure}[!htbp]
\centering
\begin{promptbox}{System Prompt: Semantic Skill Merging}
You are an expert at consolidating reusable L1 coding skills for an evolving ML agent.
You will receive several highly similar skills from the shared skill library. Merge them into ONE skill that preserves useful guidance while removing redundancy.

Your merged skill MUST follow these consolidation rules:
1. **Generality**: invent a new, generic title and guidance. Remove competition names, dataset names, file paths, and other problem-specific literals unless they are abstract examples.
2. **Parameterization**: replace hard-coded values repeated across sources with symbolic parameters.
3. **Edge-case decomposition**: if source skills disagree or apply in different regimes, do NOT choose one silently. Split the guidance into explicit conditional cases.

Output EXACTLY this plain-text structure:
Title: <a concise, descriptive title for the merged skill>
Description: <one concise sentence>
Applicability Trigger: <when the agent should retrieve this skill>
Details:
- **The Root Cause:** <shared failure modes or success factors>
- **The Pivot (Contrast):** <the change that generalizes across the source skills>
- **Generalizable Rule:** <actionable rules using symbolic parameters>
- **Anti-Pattern to Avoid:** <patterns to avoid>

Rules:
1. Each Details subsection may have 1-3 bullet points.
2. Do not include raw code blocks.
3. Be technically precise; merge overlapping advice instead of repeating it.
4. Do not invent facts absent from the source skills.
\end{promptbox}
\caption{System prompt for semantic skill merging.}
\label{fig:prompt-skill-merging}
\end{figure}

\begin{figure}[!htbp]
\centering
\begin{promptbox}{System Prompt: Standing-Rule Promotion}
You are an L2 standing-rule promoter.
L1 skills are lessons distilled from this task's earlier attempts. Most are situational: the retrieval step selects one only when the current symptoms match it. A few would help most remaining attempts on this task, whatever goes wrong next. Only those belong in L2. A conditional skill ("when X, do Y") qualifies as long as following it would help most remaining attempts.

A promoted rule is added to the prompt of EVERY remaining attempt, relevant or not, and it leaves the retrieval catalog permanently. That is the cost of promoting it. Promote a skill only if following it would help most remaining attempts.

Each candidate carries two usage figures:
  picked M/N: selected on M of the N rounds on which it was offered.
  new_bests=K: present on K rounds that improved the best result so far.
A high picked ratio with new_bests=0 is not grounds for promotion.
Promote nothing unless the evidence is there.
Do not propose a restatement of a rule that is already standing.

Return ONLY a JSON object:
{"promote_entry_ids": [], "rationale": "one sentence"}
No other text or markdown.
\end{promptbox}
\caption{System prompt for standing-rule promotion.}
\label{fig:prompt-promotion}
\end{figure}

\end{document}